\documentclass[lettersize,journal]{IEEEtran}
\usepackage{amsmath,amsfonts}
\usepackage{algorithm}
\usepackage{array}
\usepackage[caption=false,font=normalsize,labelfont=sf,textfont=sf]{subfig}
\usepackage{textcomp}
\usepackage{stfloats}
\usepackage{url}
\usepackage{verbatim}
\usepackage{graphicx}
\usepackage{cite}
\usepackage{amssymb}
\usepackage{algorithmicx,algorithm}
\usepackage{upgreek}
\usepackage{amsfonts,amssymb} 
\usepackage{caption}
\usepackage{diagbox} 
\usepackage{booktabs}
\usepackage{xcolor} 
\usepackage{multirow}
\usepackage{makecell}
\begin{document}

\title{Normality Prior Guided Multi-Semantic Fusion Network for Unsupervised Image Anomaly Detection}

\author{Muhao Xu, Xueying Zhou, Xizhan Gao, Weiye Song, Guang Feng, and Sijie Niu,\IEEEmembership{Member, IEEE}

\thanks{This work is supported by the National Natural Science Foundation of China under Grant No. 62471202, 62302191, Development Program Project of Youth Innovation Team of Institutions of Higher Learning in Shandong Province, Shandong Provincial Key Medical and Health Laboratory of Pediatric Cancer Precision Radiotherapy (Shandong Cancer Hospital), the Natural Science Foundation of Shandong Province, China, under Grant No. ZR2023QF001, Development Program Project of Youth Innovation Team of Institutions of Higher Learning in Shandong Province under Grant No. 2023KJ315, Young Talent of Lifting engineering for Science and Technology in Shandong, China, under Grant No. SDAST2024QTA014}
\thanks{Muhao Xu and Xueying Zhou contributed to the work equally and should be considered as co-first authors.
Muhao Xu, Xueying Zhou, Xizhan Gao, Guang Feng and Sijie Niu are with the Shandong Key Laboratory of Ubiquitous Intelligent Computing, University of Jinan, Jinan 250022, China.  Weiye Song and Muhao Xu are with the Department of Mechanical Engineering, Shandong University, Ji Nan, China. (Corresponding authors: Sijie Niu; Guang Feng, e-mail:sjniu@hotmail.com)} 
}

\markboth{Journal of \LaTeX\ Class Files,~Vol.~14, No.~8, August~2021}%
{Shell \MakeLowercase{\textit{et al.}}: A Sample Article Using IEEEtran.cls for IEEE Journals}


\maketitle

\begin{abstract}
Recently, detecting logical anomalies is becoming a more challenging task compared to detecting structural ones. Existing encoder–decoder based methods typically compress inputs into low-dimensional bottlenecks on the assumption that the compression process can effectively suppress the transmission of logical anomalies to the decoder. However, logical anomalies present a particular difficulty because, while their local features often resemble normal semantics, their global semantics deviate significantly from normal patterns. Thanks to the generalisation capabilities inherent in neural networks, these abnormal semantic features can propagate through low-dimensional bottlenecks. This ultimately allows the decoder to reconstruct anomalous images with misleading fidelity. To tackle the above challenge, we propose a novel normality prior guided multi-semantic fusion network for unsupervised anomaly detection. Instead of feeding the compressed bottlenecks to the decoder directly, we introduce the multi-semantic features of normal samples into the reconstruction process. To this end, we first extract abstract global semantics of normal cases by a pre-trained vision-language network, then the learnable semantic codebooks are constructed to store representative feature vectors of normal samples by vector quantisation. Finally, the above multi-semantic features are fused and employed as input to the decoder to guide the reconstruction of anomalies to approximate normality. Extensive experiments are conducted to validate the effectiveness of our proposed method, and it achieves the SOTA performance on the MVTec LOCO AD dataset with improvements of 5.7\% in pixel-sPRO and 2.6\% in image-AUROC. The source code is available at https://github.com/Xmh-L/NPGMF.
\end{abstract}


	\section{Introduction}
Unsupervised image anomaly detection \cite{li2022unsupervised,tao2023vitalnet} is usually desirable to train the models solely on anomaly-free image data for identifying and localizing the anomalies during inference, which has attracted increasing attention in the research community and has been widely applied in fields of computer vision, such as medical disease diagnosis \cite{chen2022deep,xia2022gan}, industrial defect detection \cite{yang2023cloud,cai2023itran}, and road safety monitoring \cite{li2014crowded,10836187}. 

\begin{figure}[tbp]
  \centering
  \includegraphics[width=0.48\textwidth]{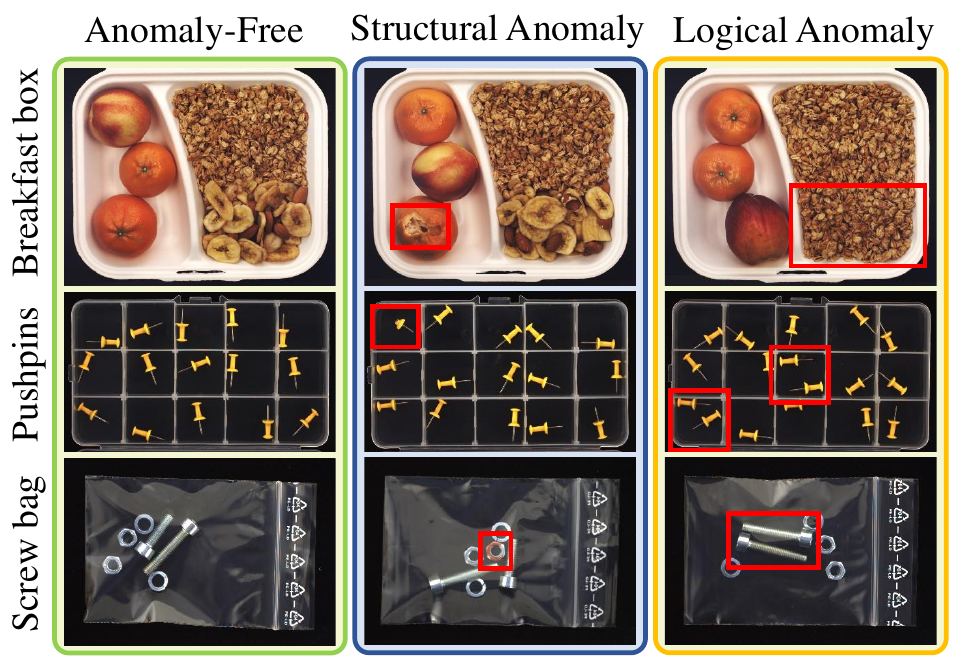}
  \caption{Samples of different types of anomalies, including normal images (left), structural (middle) and logical anomalies (right). The structural anomalies introduce novel local structures (i.e., damaged tangerine, broken pushpin, and stained screw), and the logical anomalies violate logical constraints of the normal images (i.e., the missing banana chips and almonds, too many pushpins in each compartment, and the incorrect number of long and short screws).} 
   \label{IN}
\vspace{-0.3cm}
\end{figure}

Most existing unsupervised image anomaly detection methods \cite{roth2022towards,liu2023anomaly,xu2022discriminative} share a common training strategy, which involves training a separate detection model for each category of images. This strategy is employed to mitigate interference from inter-class information during both training and inference stages and successfully detect anomalies by finding the deviations in their appearance and behavior. These deviations are mainly divided into structural anomalies and logical anomalies. Structural anomalies refer to entirely new local structures compared to normal cases, and the definition of logical anomalies is the contents that violate the underlying logical or geometric constraints of the normal samples. To illustrate these two different types of anomalies, some examples from the MVTec LOCO AD \cite{bergmann2022beyond} dataset are presented in Figure \ref{IN}.

A wide range of methods have been proposed in recent years for unsupervised anomaly detection, including the representation memory-bank framework \cite{xu2022discriminative,reiss2021panda,park2020learning,xing2023visual,wu2022self} and the encoder-decoder framework \cite{bergmann2022beyond,bergmann2019mvtec,deng2022anomaly,wang2018generative,guo2023template,dong2009pointwise}. 
Among them, PatchCore \cite{roth2022towards}, a representative approach of the memory-bank paradigm, has achieved state-of-the-art (SOTA) performance on datasets dominated by structural anomalies (e.g., MVTec AD \cite{bergmann2019mvtec})without requiring fine-tuning or model retraining. 
During training, it stores the local features of normal images extracted by a pre-trained network, and detects anomalies by measuring the discrepancy between the representations of test samples and those stored in memory. 
However, because it lacks the ability to model global contextual information, PatchCore and similar approaches struggle to effectively detect and localize logical anomalies. 
Thus, detecting logical anomalies is widely regarded as a more challenging task in the field of anomaly detection \cite{bergmann2022beyond}.

The encoder-decoder architecture is capable of capturing global context information by compressing input images into low-dimensional bottlenecks, making it particularly effective for identifying logical anomalies.
For example, RD \cite{deng2022anomaly} introduces a one-class bottleneck embedding module that projects high-dimensional features into a compact latent subspace of normal samples. 
GCAD \cite{bergmann2022beyond} proposes to learn globally consistent representations of normal data by compressing their high-dimensional embeddings into low-dimensional bottlenecks. 
These methods are motivated by the insight that the bottleneck representation enforces a strong global constraint, which tends to suppress logical anomalies during reconstruction, thereby amplifying the representation discrepancy between normal and abnormal samples. Recently, VLMs have been introduced into anomaly detection due to their powerful capability in capturing high-level semantic and contextual information.
LogicAD \cite{jin2025logicad} represents a significant step forward in explainable anomaly detection by harnessing the power of alignment pre-trained VLMs (AVLMs) for text-based feature extraction and logical reasoning.
These models leverage the semantic alignment between image and text modalities to detect anomalies that are semantically inconsistent with learned normal patterns.

Though prior works provide effective solutions to the task of logical anomaly detection, several important issues remain unexplored.
First, the assumption that logical anomalies cannot be reproduced during inference does not hold at all time \cite{zhou2021memorizing}.  However, owing to the generalisation capacity of neural networks, particularly when handling data with complex semantics, some anomalous features might still be captured, thereby undermining this assumption. Thus, abnormal semantics probably still play a leading role in the bottlenecks that are fed into the decoder, decreasing the representation discrepancy between the encoder and decoder on anomalies.  
Second, the decoder attempts to reconstruct high-dimensional representations from the low-dimensional bottleneck. However, the reconstructions are always ambiguous and inaccurate due to the lack of different level features in low-dimension bottlenecks. To overcome that, RD \cite{deng2022anomaly} introduced multi-level representations extracted from the pre-trained network in the reconstruction process. However, this processing increases the risk of anomaly feature leakage at the inference stage, which refers to the phenomenon where the reconstruction process inadvertently transmits certain anomalous features from the multi-level representations to the decoder. This leakage diminishes the discrepancy between the reconstructed output and the original anomalous input, adversely impacting the detection performance.

To overcome the abovementioned problems, we propose a novel normality prior guided multi-semantic fusion network for unsupervised anomaly detection, which can detect and localize both structural and logical anomalies well. 
For logical anomaly detection, we first introduce a pre-trained encoder to extract different level features of input images and compress high-level features into low-dimensional bottlenecks. To degrade the impact of abnormal semantics on reproduction, we then introduce multi-semantic features of normal samples into the decoder. Specifically, we employ a pre-trained vision-language network, CLIP \cite{radford2021learning}, to capture the abstract global context of the image for the given caption annotation, and construct the learnable discrete codebook through vector quantization to obtain the representative different level features of normal data. By utilizing the normality prior, we guide the fusion of multi-semantic features in such a way that the decoder prioritizes normal semantics. This strategy aids in suppressing the influence of anomalous features, thereby enhancing the representation discrepancy between normal and abnormal instances.
Additionally, to balance the ability of our proposed method to tackle structural anomalies, following previous works \cite{bergmann2022beyond}, a simple structural anomaly detection module is constructed based on the representation memory-bank framework. Extensive experiments are conducted on two publicly-available anomaly detection datasets, and our proposed method achieves the state-of-the-arts performance on MVTec LOCO AD dataset (e.g., +5.7\% pixel-sPRO and +2.6\% image-AUROC) as well as competitive performance on MVTec AD. In conclusion, the main contributions of our proposed method can be summarized as follows: 
\begin{enumerate}
	\item We propose a novel multi-semantic fusion network for unsupervised anomaly detection. By fusing abstract global context features, multi-level features, and low-dimensional bottlenecks, our method effectively increases the discrepancy between abnormal inputs and their reconstructions, thereby significantly enhancing anomaly discrimination.
	\item We construct learnable normal multi-level codebooks through vector quantization techniques to capture representative features from normal data. During testing, these normal codebooks are utilized in the decoder to prevent abnormal feature leakage, ensuring more robust reconstruction of normal semantics.
	\item We integrate a structural anomaly detection module based on the representation memory-bank framework. This additional module complements our fusion network and contributes to superior localization and detection performance across a variety of anomaly types.
\end{enumerate}


\section{Related Work}
This section briefly reviews previous efforts on unsupervised anomaly detection and vision-language pre-trianed networks.

\textbf{Structural Anomaly Detection.}
Structural anomalies are defined as entirely new local structures compared to normal cases. Methods that detect anomalies in units of local representations have achieved excellent performance in structural anomaly detection. In the training phase, these methods introduce the pre-trained networks to extract representations (i.e., feature vectors or feature maps) of images and model the representation distribution of normal data. In the inference phase, the anomalies can be captured by the big difference between representations of test samples and normal cases. Initially, SPADE \cite{cohen2020sub} proposed to store the representations of normal samples in a memory bank, and detected anomalies by representation difference between test images and the stored ones. To make the model more adaptive to the target distribution, some methods \cite{reiss2021panda} re-trained the pre-trained networks in the training stage. Additionally, teacher-student framework \cite{bergmann2020uninformed,wang2021student,salehi2021multiresolution} is one of the most classic representation-based anomaly detection methods. For these methods, the student networks are trained by decreasing the difference between their outputs and those of the pre-trained teacher network, thus the unseen anomalies can be detected due to the huge discrepancy between the outputs of the teacher network and those of the student network. Due to the use of local representations rather than global context features, representation-based methods are sensitive to structural anomalies but insensitive to logical ones.

\textbf{Logical Anomaly Detection.}
Logical anomalies refer to deviations that violate the underlying logical or geometric constraints inherent in the training data. 
It is generally believed that compressing and reconstructing normal samples helps to model these global constraints, making it difficult for logical anomalies—which deviate from such constraints—to be accurately reconstructed. 
As a result, encoder-decoder architectures are considered to be more effective in capturing logical anomalies. 
These models are typically trained to minimize the reconstruction error on normal images and identify anomalies by measuring significant discrepancies between the input and the reconstructed output. 
To further amplify the reconstruction error between normal and abnormal samples, recent approaches—such as generative adversarial networks (GAN)-based methods \cite{akcay2019ganomaly}, variational autoencoders (VAE) \cite{liu2020towards,sun2020adversarial}, and memory network-based techniques \cite{xu2022discriminative,roth2022towards,reiss2021panda}—have introduced additional constraints in the latent space of the reconstruction model. 
These strategies have demonstrated superior performance compared to conventional vanilla autoencoders (AE) \cite{deng2022anomaly,bergmann2022beyond,wang2018generative,zhou2022pull}.
More recently, RD \cite{deng2022anomaly} and GCAD \cite{bergmann2022beyond} have incorporated teacher-student learning frameworks into encoder-decoder models, detecting anomalies based on discrepancies between reconstructed representations and those derived from pre-trained networks. 
Despite these advancements, reconstruction-based methods still suffer from high false positive rates due to their inherent ambiguity and imperfect reconstruction quality. 
To address this, LogicAD \cite{jin2025logicad} introduces a novel paradigm by leveraging alignment-based vision-language models (AVLMs) for text-driven feature extraction and logical reasoning, significantly enhancing explainability in anomaly detection. 
However, its limited ability to handle low-level structural anomalies highlights the need for further refinement to ensure robust performance across both logical and structural anomaly categories in industrial applications.
In summary, while reconstruction-based frameworks and recent vision-language models have advanced the detection of logical anomalies, balancing semantic reasoning and fine-grained structural perception remains an open challenge in building unified, generalizable anomaly detection systems.

\textbf{Vector Quantised-Variational AutoEncoder.} Vector Quantized Variational Autoencoder \cite{van2017neural} integrates variational autoencoders with vector quantization techniques primarily for generative modeling and data compression tasks. The implementation of the VQ-VAE model maps the input data into a latent space through an encoder, which subsequently uses a vector quantization technique to quantize the vectors in this latent space into a set of discrete fixed vectors, i.e., the codebook. Finally, the decoder maps these discrete vectors back into the original latent space, which is then decoded into the reconstructed input data. In generative modeling \cite{rombach2022high,peng2021generating}, it produces high quality data samples that are similar to the training data by learning their latent representations. Meanwhile, in data compression \cite{garbacea2019low,guo2023msmc}, VQ-VAE employs vector quantization techniques to represent high-dimensional data as compact, discrete forms, significantly reducing storage space while retaining essential information. Recently, this approach has also been used for anomaly detection \cite{pinaya2022unsupervised,pinaya2022fast,wang2023intrusion}.

\begin{figure}[tbp]
  \centering
  \includegraphics[width=0.48\textwidth]{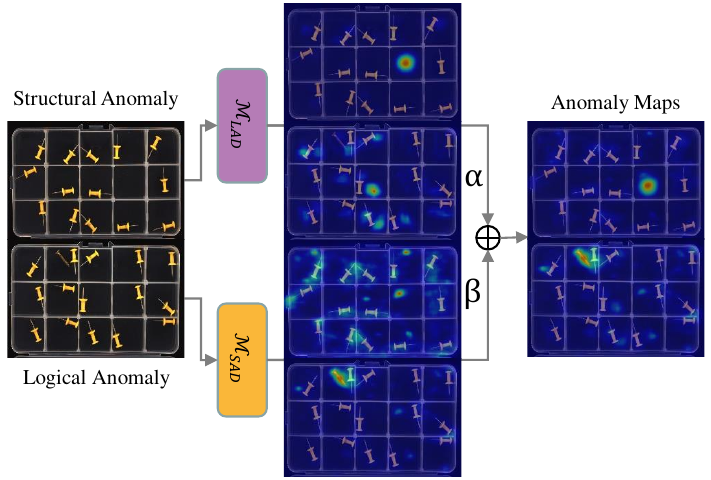}
  \caption{Overview architecture of our proposed approach.} 
   \label{A1}
\vspace{-0.2cm}
\end{figure}

\textbf{Vision-Language Pre-trained network.}

Recently, vision-language pre-trained models \cite{zhou2020unified,radford2021learning,li2022label2label,tang2021clip4caption,ou2021multimodal} have garnered significant attention in transfer learning and demonstrated remarkable success across a wide range of downstream tasks. 
For example, VirTex \cite{desai2021virtex} and ICMLM \cite{li2022label2label} exploit semantically rich captions to learn high-quality visual representations using only a fraction of the data required by traditional methods such as ImageNet, while achieving comparable or even superior performance. 
Despite these advances, most existing methods \cite{tang2021clip4caption,zhou2020unified,xu2022simple,desai2021virtex,li2022label2label} still rely on fine-tuning pre-trained models using target-domain data, which constrains their generalizability and limits real-world applicability. 
To address this limitation, CLIP \cite{radford2021learning} was introduced as a more robust vision-language model capable of zero-shot transfer to diverse downstream tasks without task-specific training. 
CLIP has since been widely applied to areas such as video-text retrieval \cite{ma2022x,liu2021hit}, image generation \cite{wang2022clip}, classification \cite{islam2022long}, and segmentation \cite{wang2022cris,park2022per}, showcasing its versatility and generalization capabilities. 
However, its potential in anomaly detection remains largely underexplored, despite the task's inherent semantic complexity and practical significance. 
In summary, the integration of vision-language models such as CLIP into anomaly detection presents a promising direction, with the potential to improve semantic reasoning and reduce reliance on domain-specific annotations.

  \begin{figure*}
  \centering
  \includegraphics[width=\textwidth]{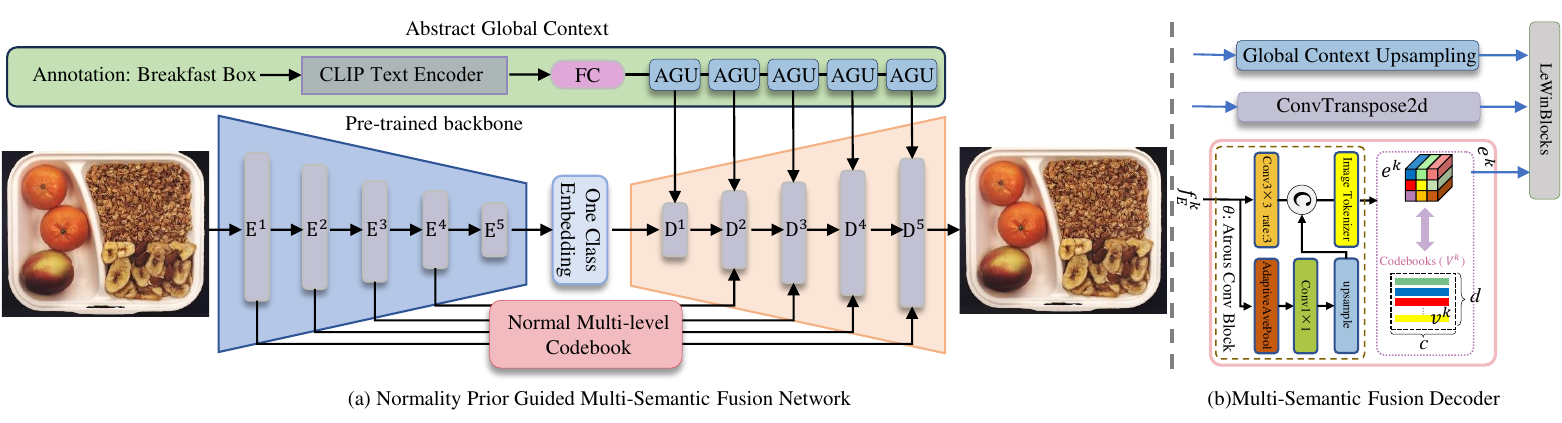}
   \caption{Overview architecture of proposed Normality Prior Guided Multi-Semantic Fusion Network. (a) Normality Prior Guided Multi-Semantic Fusion Network consists of one pre-trained backbone encoder, normal multi-level codebook, abstract global context, and a multi-semantic restoration decoder. Image features are extracted from test images using the encoder pre-trained on ImageNet. The category text for the image is encoded using the pre-trained CLIP text encoder, which provides global semantic features. During the learning process of the student encoder, the image features, global semantic features, and multilevel features from the normal image are fused together to enhance the accuracy of normal image reconstruction. The anomaly is detected by observing a difference between the anomaly-free recovered features and those present in the input. The details of the proposed method are introduced in Sec. \ref{load}.(b) Detailed network design of multi-semantic restoration decoder.}
  \label{MOAD}
\vspace{-0.3cm}
\end{figure*} 

\section{Proposed Method}

Our proposed approach is comprised of two components: the Logical Anomaly Detection Module $\mathcal{M}_{LAD}$ and the Structural Anomaly Detection Module $\mathcal{M}_{SAD}$, which are primarily responsible for detecting corresponding types of anomalies. The Logical Anomaly Detection Module and the Structural Anomaly Detection Module are trained separately to compute anomaly scores. In the inference phase, the anomaly scores obtained from the two modules are combined to get the final anomaly scores. The overall framework is shown in Fig.\ref{A1}. The problem formulation of our proposed method is as follows.

Given datasets $\mathcal{D}_{train}=\left\{ x_{1},x_{2},...,x_{n} \right\} $ with $n$ anomaly-free images $ x \in \mathbb{R}^{H \times W \times C} $ and $ \mathcal{D}_{test}=\left\{ (x_{1}^{t},y_{1}),(x_{2}^{t},y_{2}),...,(x_{n_t}^{t},y_{n_t}) \right\} $ containing $n_t$ normal and abnormal images and their labels $y$, where $y \in \{0,1\}$, $0$ indicates normal and $1$ indicates abnormal. The purpose of our method is to detect anomalies in $\mathcal{D}_{Test}$ though fusing the anomaly scores calculated by the modules $\mathcal{M}_{LAD}$ and $\mathcal{M}_{SAD}$, which are trained on the $\mathcal{D}_{Train}$.

\subsection{Logical Anomaly Detection Module}
\label{load}
The Logical Anomaly Detection Module $\mathcal{M}_{LAD}$ consists of two critical parts: the pre-trained encoder $E$ and the multi-semantic fusion decoder $D$, and the architecture of our $\mathcal{M}_{LAD}$ are shown in Fig. \ref{MOAD}.

\subsubsection{Overview}

Following previous works \cite{deng2022anomaly,bergmann2022beyond}, a frozen encoder $E$ pre-trained on ImageNet is introduced to extract different level features $f_E^k \in \mathbb{R}^{C^k \times H^k \times W^k}$ of input images through correspondding encoding blocks $E^k$, where $k \in \{1, 2, 3, 4, 5\}$ ($E^5$ is the last convolution block of the encoder network). 
To capture the global constraint of input images, the output ($f^5_E$) of $E^5$ are fed into the One Class Embedding (OCE) module to obtain the one-class features. The OCE module consists of two convolutional layers with strides of 2. Then, a trainable decoder is constructed to reproduce different level features $f_D^k \in \mathbb{R}^{C^k \times H^k \times W^k}$ from the one-class features by the corresponding decoding blocks $D^k$.

\begin{figure}[tbp]
  \centering
  \includegraphics[width=0.48\textwidth]{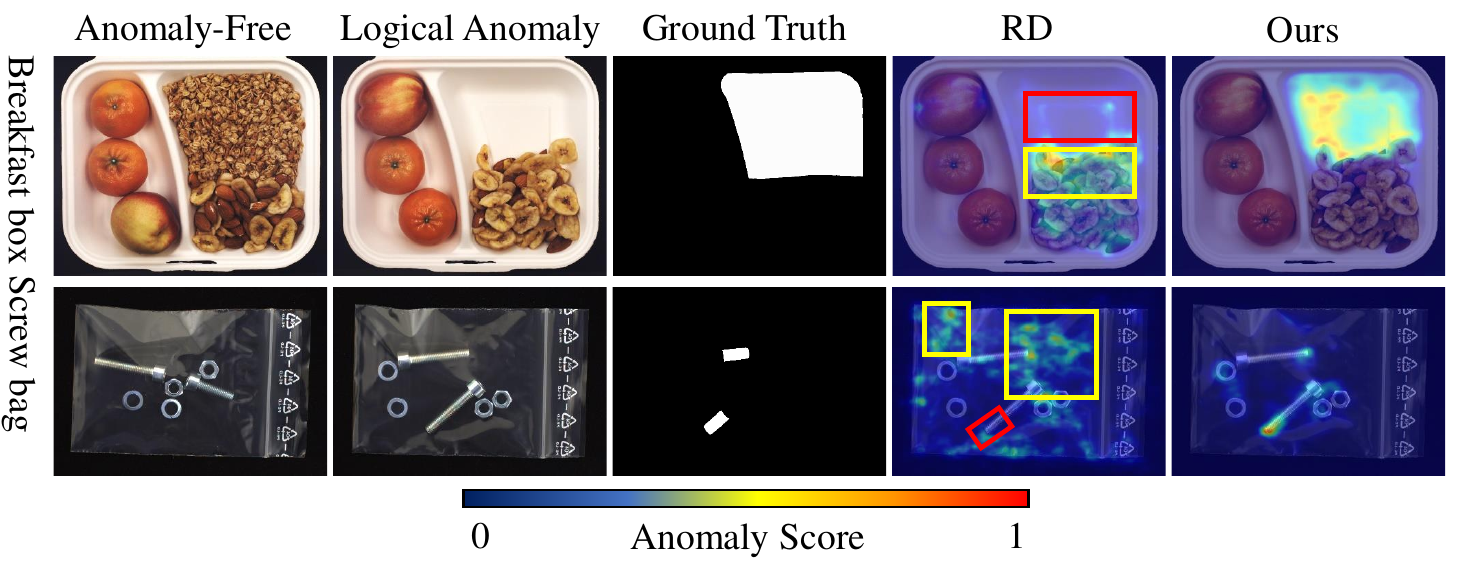}
  \caption{Visualization of logical anomaly localization results of our proposed methods and RD \cite{deng2022anomaly}.} 
   \label{OV}
\vspace{-0.2cm}
\end{figure}

A paradigm leveraged by previous methods to detect anomalies is minimizing the differences between representations obtained by encoder and decoder, and detecting anomalies through encoder-decoder representation discrepancy.
Although the above paradigm has a certain effect on logical anomaly detection, two shortfalls restrict the improvement of which. On the one hand, abnormal semantics inevitably play a leading role in the one-class features due to the generalization ability of the deep neural network, which reduces the sensitivity of the above paradigm to logical anomalies. On the other hand, it is challenging to reconstruct high-dimensional features from low-dimensional ones, leading to false positives (i.e., normal regions with high anomaly scores). To overcome that, RD proposed to compress multi-level features and utilized them to reconstruct high-dimensional features. However, these one-class features extracted from the pre-trained encoder introduce anomalies in the reconstruction process, causing high false negatives (i.e., abnormal regions with low anomaly scores). As shown in Figure \ref{OV}, RD cannot well localize logical anomalies (e.g., red boxes) and reports high false positives (e.g., yellow boxes).  

To address the above issues, we construct a novel multi-semantic fusion decoder. By introducing multi-semantic features of normal samples in the process of reconstruction, the logical anomalies can be well captured by large reconstruction errors. Firstly, we obtain abstract global context of normal samples by a vision-language pre-trained network, which will not be disturbed by abnormal inputs during inference (Sec. \ref{CGS}). Secondly, learnable semantic codebooks, trained to memory critical different level features of normal samples, are constructed by vector quantisation. In the inference phase, the memorized features are introduced to reconstruction, which is beneficial to improving reconstruction quality without introducing anomalies (Sec. \ref{CMSC}). Finally, features with different semantics are fused and are used to guide the reproduction of different level features, making them approximate normality. (Sec. \ref{MSFF}). Therefore, as reported in Figure \ref{OV}, our proposed method not only accurately localizes logical anomalies, but also reports few false positives. Furthermore, the details of our multi-semantic fusion decoder are introduced in the following subsections.

\begin{figure}[tbp]
  \centering
  \includegraphics[width=0.48\textwidth]{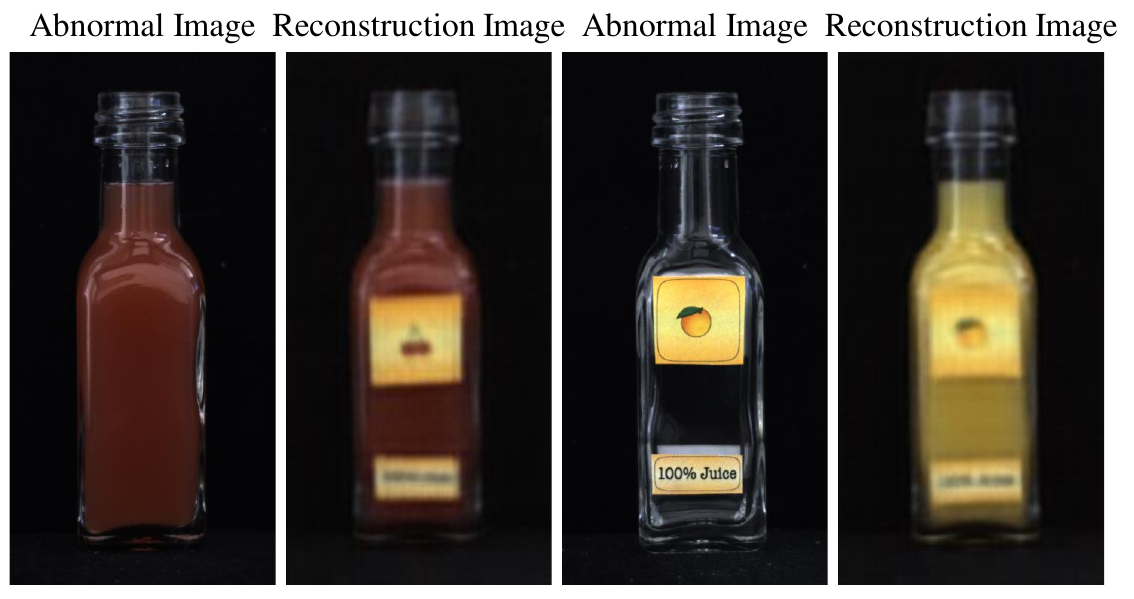}
  \caption{Visualization of the reconstructed results.} 
   \label{CY}
\end{figure}

\subsubsection{Abstract Global Context}
\label{CGS}

CLIP \cite{radford2021learning}, proposed by OpenAI, offers a pre-trained vision-language model that can be used to generate the corresponding abstract global context of images for the given text inputs.  

For abnormal inputs, abnormal semantics probably still play a leading role in the one-class features extracted by the encoder. To weaken the contribution of the abnormal semantics in the reconstruction process, an essential way is to introduce abstract global context features of normal data.
Fortunately, the pre-trained CLIP can capture the abstract context features of the images.
The category information for each image is actually known, while the annotations we fed into CLIP are category names (e.g., `Breakfast Box') given by the training dataset. Thus, no supervision of anomalies is introduced in the training phase. 
Taking advantage of it, the abstract global context features of images are obtained by feeding their caption annotation (we use category names in this paper) into the pre-trained text encoder of CLIP.

To adapt the features extracted by the CLIP text encoder to the current dataset, we use a fully-connected layer to re-learn the features. 
To be specific, the parameters of the pre-trained CLIP text encoder are frozen during the training and inference stages. During the training stage, the output features of the CLIP text encoder are fed into a fully connected layer (FC). We adapt the FC layer to the target data by updating its parameters, enabling it to capture the abstract global context features of normal images.
Thus, the discrepancies between abnormal inputs and their reconstruction are boosted due to the introduction of the normal semantics in the inference phase. To incorporate the extracted abstract global context features into the reconstruction process, we employ a upsampling to progressively expand them to match different spatial dimensions and concatenate them with the image features of the corresponding layers.

\subsubsection{Normal Multi-level Codebook}
\label{CMSC}

For the sake of bringing multi-level features of normal samples to the test phase, the learnable discrete codebooks $V^k \in \mathbb{R}^{d \times\frac{C^k}{4}}$, containing $d$ entries $v^k \in \mathbb{R}^{\frac{C^k}{4}}$  ($d$ is set to 16 for different level codebooks), are constructed through vector quantisation (VQ). To be specific, the features with multi-scale contextual information is obtained by feeding the features $f_E^k(:,h^k,w^k)$  into the atrous convolution network $\theta$, where $h^k \in \{0,...,H^k\}$ and $w^k \in \{0,...,W^k\}$ is the coordinate of feature vectors in spatial dimension. The flowchart of the atrous convolution block is reported in Figure \ref{MOAD}. Next, these feature vectors are represented by the entries stored in the corresponding codebooks, the process can be written as:

\begin{flalign}
\begin{aligned}
e^k(:,h^k,w^k)& = \underset{v^k \in V^k}{\arg\min}\|\theta (f_E^k(:,h^k,w^k))-v^k\|_2.
\label{VQVAE}
\end{aligned}
\end{flalign}

By repeating the above operation, the final output features $e^k \in \mathbb{R}^{\frac{C^k}{4} \times H^k \times W^k}$ are obtained.

During model training, the codebooks are trained to store semantic-rich feature vectors of normal samples. However, the quantisation operation in Eq.\ref{VQVAE} is non-differentiable. To address this issue, a straight-through gradient estimator \cite{van2017neural} is utilized to achieve backpropagation by simply copying the gradients from the inputs to the outputs of quantisation operation. Thus, the optimization objects of the proposed codebooks are given as:
\begin{equation}
\begin{aligned}
\mathcal {L}_{vq}&=\lVert sg[f_E^k] - e^k \rVert^2_2+\lVert f_E^k- sg[e^k]\rVert^2_2,
\end{aligned}
\end{equation}
where $\mathrm{sg}[\cdot]$ represents the stop-gradient operator. In test time, the different level features of normal samples stored in the codebooks are introduced into the reconstruction process through Eq. \ref{VQVAE}, so that the anomalies can be accurately reconstructed into normal cases.

\subsubsection{Multi-Semantic Feature Fusion}
\label{MSFF}

To introduce the global context information of normal samples,
we enhance the reconstruction features by inputting $ f_E^5 $ into the One Class Embedding (OCE) module. Then we concatenate the reconstruction and global context features in the channel dimension and perform feature fusion by feeding them into LeWinBlocks. Furthermore, we concatenate the reconstruction features $f_D^{k-1}$, the global context features and vector quantization features  $e^k$ and perform feature fusion through the LeWinBlocks module, then send to the corresponding decoding block $D^k$. After iterations, the reconstruction images $x' \in \mathbb{R}^{H \times W \times C}$ are obtained.

\subsubsection{Loss Function}

In order to improve the ability of the decoder to reconstruct normal images, feature reconstruction loss $\mathcal {L}_{cos}$ and image reconstruction loss $\mathcal {L}_{mse}$ are constructed through cosine similarity and mean square error (MSE), which are given as follows:
\begin{equation}
\begin{aligned}
M^k(h^k,w^k)&=1-cosine(f_E^k(:,h^k,w^k), f_D^k(:,h^k,w^k) ),
\end{aligned}
\end{equation}
\begin{equation}
\begin{aligned}
\mathcal {L}_{cos}&=\sum\limits^{4}_{k=1}(\frac{1}{H^k\times W^k}\sum\limits^{H^k}_{h^k=1}\sum\limits^{W^k}_{w^k=1}{M^k(h^k,w^k))},
\end{aligned}
\end{equation}
\begin{equation}
\begin{aligned}
\mathcal {L}_{mse}&=MSE(x,x').
\end{aligned}
\end{equation}
Thus, the total loss can be written as:
\begin{equation}
\begin{aligned}
\mathcal {L}_{total}&= \lambda_1 \mathcal {L}_{cos}+ \lambda_2 \mathcal {L}_{mse}+ \lambda_3 \mathcal {L}_{vq}.
\end{aligned}
\end{equation}

In the inference phase, the test abnormal samples are fed into the encoder to extract the features with abnormality, while the multi-semantic fusion decoder is capable of reconstructing them into anomaly-free ones, as shown in Figure \ref{CY}. In other words, the multi-semantic fusion decoder $D$ generates discrepant representations from the pre-trained encoder $E$ when the query is anomalous. Thus, we obtain a set of anomaly score maps $M^k$ of different levels, where each pixel value in the map represents the likelihood that pixels are classified as logical anomalies. To localize anomalies in a test sample, we up-sample $M^k$ to the image size. Let $\varphi$ denote the bilinear up-sampling operation used in this work. The logical anomaly score map $\mathcal{A}_{log} \in \mathbb{R}^{H \times W}$ is formulated as: 
\begin{align}
    \mathcal A_{log}=\sum\limits^{4}_{k=1}{\varphi(M^{k})}.
\label{alog}
\end{align}

\begin{table*}

  \caption{Quantitative results on MVTec LOCO AD dataset for anomaly localization, as measured on pixel-sPRO. The best results are marked in bold.}
\centering 

\begin{tabular}{l|cccccc}
    \toprule
   \rule{0pt}{10pt}  Method & Breakfast Box & Screw Bag & Pushpins & Splicing Connectors & Juice Bottle & Mean \\
    \midrule
    VM \cite{steger2018machine}   & 0.168 & 0.253 & 0.254 & 0.125 & 0.325 & 0.225 \\
    f-AnoGAN \cite{schlegl2019f} & 0.223 & 0.348 & 0.336 & 0.195 & 0.569 & 0.334 \\
    MNAD  \cite{park2020learning}  & 0.080  & 0.344 & 0.357 & 0.442 & 0.472 & 0.339 \\
    AE  \cite{an2015variational}   & 0.189 & 0.289 & 0.327 & 0.479 & 0.605 & 0.378 \\
    VAE \cite {bergmann2018improving}  & 0.165 & 0.302 & 0.311 & 0.496 & 0.636 & 0.382 \\
    SPADE  \cite{cohen2020sub}& 0.372 & 0.331 & 0.234 & 0.516 & 0.804 & 0.451 \\
    S–T \cite{bergmann2020uninformed}   & 0.496 & 0.602 & 0.523 & 0.698 & 0.811 & 0.626 \\
RD  \cite{deng2022anomaly}  & 0.560 & 0.535 & 0.577 & 0.701 & 0.837 & 0.642 \\

PaDiM \cite{defard2021padim}  &0.509 & 0.461 & 0.295 &0.467 &0.779 & 0.502 \\
Patch Core  \cite{roth2022towards}  & 0.451 &0.562  & 0.423 & 0.598 & 0.694 & 0.546 \\
    GCAD \cite{bergmann2022beyond} & 0.502 & 0.558 & 0.739 & 0.798 & \textbf{0.910} & 0.701 \\
    Ours & \textbf{0.649} & \textbf{0.666} & \textbf{0.755} & \textbf{0.822} & 0.896 & \textbf{0.758} \\
    \bottomrule
    \end{tabular}%
  \label{tab:f}%
\end{table*}%

 \begin{figure}[htbp]
  \centering
  \includegraphics[width=0.48\textwidth]{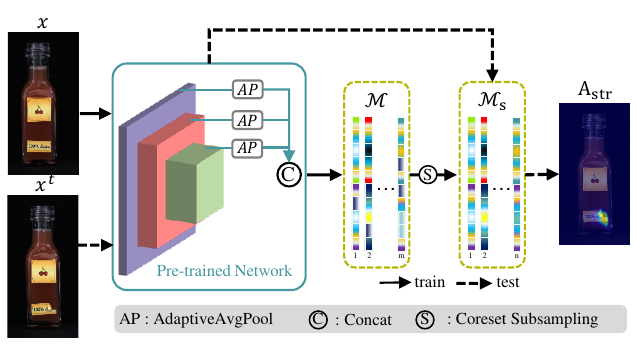}
  \caption{Flowchart of our structural anomaly detection framework.} 
   \label{SAD}
\vspace{-0.2cm}
\end{figure}

\subsection{Structural Anomaly Detection Module }
To balance the ability of the model to detect multiple anomalies, a simple representation memory-bank framework based method is constructed as the Structural Anomaly Detection Module $\mathcal{M}_{SAD}$, as illustrated in Figure. \ref{SAD}.

Firstly, we extract different level features of training data by the pre-trained network and obtain multi-level aggregation features $f_s \in \mathbb{R}^{C_s \times H_s \times W_s}$. Then, to maximize available information of normal samples at inference time, a memory bank $\mathcal{M} \in \mathbb{R}^{(H_s \times W_s \times n) \times C_s} $ is constructed to store feature vectors of training samples, which can be formulated by:
\begin{flalign}
\mathcal{M}=\bigcup_{(h_s,w_s) \in \{(h_s,w_s)| h_s \in \{0,...,H_s\}, w_s \in \{0,...,W_s\}\}}{f_s(:,h_s,w_s)}, 
\end{flalign}

where $f_s(h_s,w_s) \in \mathbb{R}^{C_s}$ represents the multi-level aggregation feature vectors. Moreover, to reduce storage and computational costs in the inference phase, we use greedy coreset subsampling \cite{dasgupta2003elementary} to select the smallest subset $\mathcal{M}_s \in \mathbb{R}^{n_s \times C_s}$ ($n_s\ll(W_s \times H_s \times n$)) that can represent the memory bank $\mathcal{M}$.

In test time, the anomalies can be captured by the large difference between feature vectors of the test samples and those of the stored ones, so the  structural anomaly score map $\mathcal{A}_{str} \in \mathbb{R}^{H \times W}$ is defined as follows:
\begin{align}
    \mathcal A_{str}=\varphi(M_s),
\label{astr}
\end{align}
\begin{align}
   M_s(h_s,w_s)= \mathop{\arg\min} \limits_{1 \le u \le s}{\|f_s(h_s,w_s)-\mathcal{M}_S(u,:)\|_1}.
\end{align}

\subsection{Anomaly Score}
At the inference stage, the image- and pixel-level anomaly scores are constructed to calculate the likelihood of the test samples and their pixels being anomalies, respectively. Specifically, for each test image $ x^t \in \mathbb{R}^{H \times W \times C} $, an anomaly score map $\mathcal{A}_{map} \in \mathbb{R}^{H \times W}$, represents pixel-level anomaly score of each position, is calculated. And the largest pixel anomaly score of each image is denoted as the anomaly score of the image. 

 To obtain the anomaly score maps, we first feed the test samples into logical and structural anomaly detection module, respectively. Then, the logical and structural anomaly score maps $\mathcal{A}_{log}$ and $\mathcal{A}_{str}$ are obtained by Eq.\ref{alog} and Eq.\ref{astr}.

Finally, the anomaly score maps are computed by merging the logical and structural anomaly score maps.

\begin{equation}
\begin{aligned}
	\mathcal A_{map}=\upalpha \frac{\mathcal A_{str}-\mu_{str}}{\sigma_{str}}+ \upbeta \frac{\mathcal A_{log}-\mu_{log}}{\sigma_{log}},
\end{aligned}
\end{equation}
where $\mu$ and $\sigma$ are the mean value and standard deviation of all pixel-level anomaly scores on the given validation dataset, and $\upalpha$ and $\upbeta$ are the weights for balancing the structural and logical anomaly scores. Eventually, gaussian filter with $\sigma = 4$ is used to smooth the anomaly score map.

\begin{figure*}

  \includegraphics[width=\textwidth]{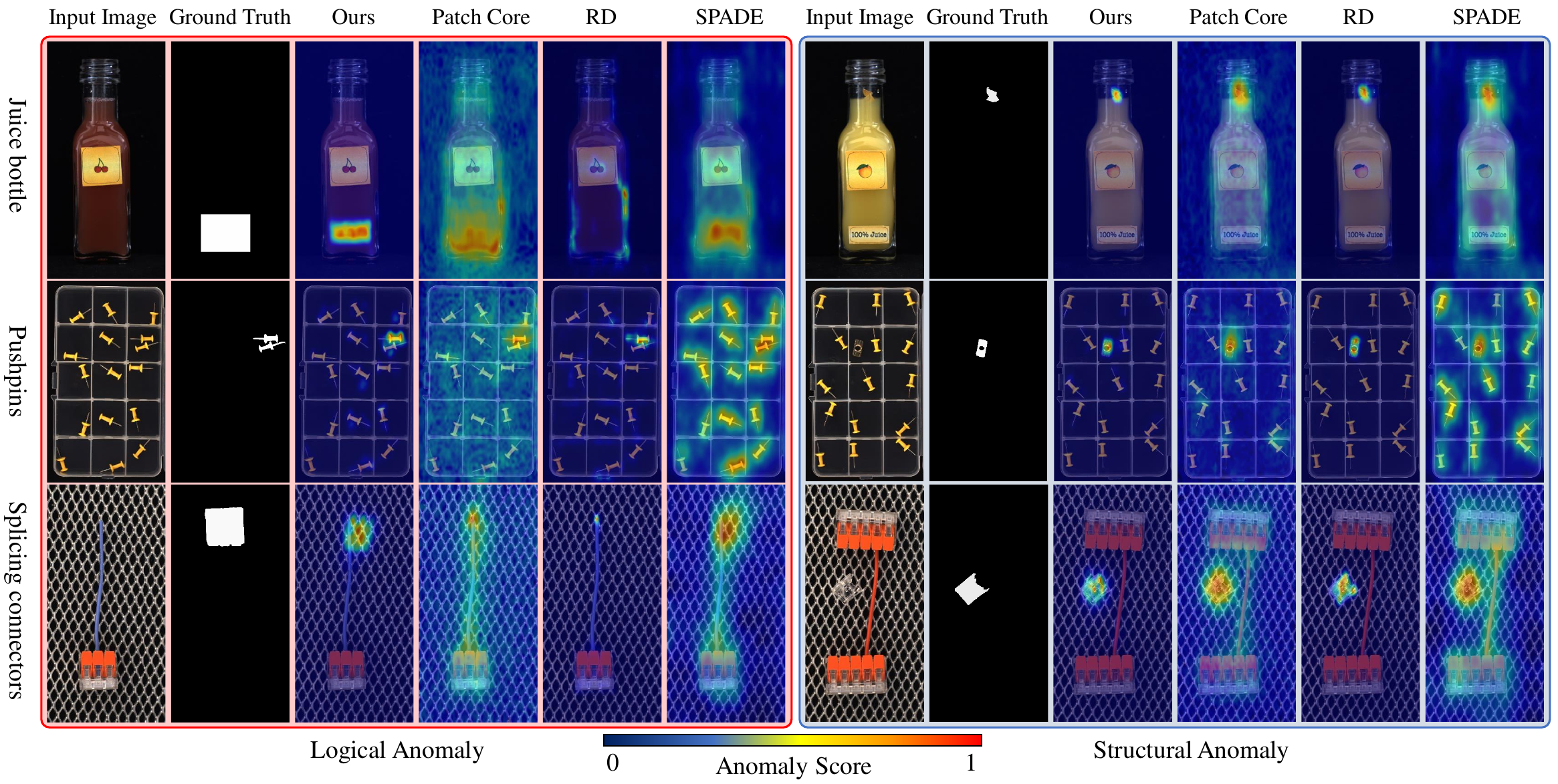}

 \caption{Visualization of anomaly localization results on different categories of MVTec LOCO AD.} 
   \label{Vision}
\end{figure*}

\section{Experiments}

To validate the effectiveness of our proposed method, the empirical evaluations on the MVTec LOCO AD and MVTec AD datasets are performed. Moreover, discussions in both parameter analysis and ablation study are implemented on the MVTec LOCO AD dataset. 
\subsection{Settings}

\textbf{MVTec LOCO AD dataset \cite{bergmann2022beyond}.} A novel dataset for structural and logical anomaly detection and localization. Structural anomalies manifest as scratches, dents, or contaminations in manufactured products. On the other hand, logical anomalies involve violations of underlying constraints or rules. Examples include an object being present in an invalid location or a required object missing altogether. These logical anomalies emphasize the importance of capturing contextual relationships and dependencies. The dataset consists of 3644 images from five different categories, inspired by real-world industrial inspection scenarios. The training set and validation set contain only anomaly-free samples, while both normal and abnormal samples are contained in the test set. 

\textbf{MVTec AD dataset \cite{bergmann2019mvtec}.}  It is one of the most challenging anomaly detection datasets mainly dominated by structural
anomalies. These anomalies may arise due to object damage, defects, deformations, or other structural issues. It contains 15 categories of real-world datasets, including 5 classes of textures and 10 classes of objects. Each category is divided into training set and test set, and there are 3629 normal training samples and 1725 normal and abnormal test samples in total.

\textbf{ BeanTech AD (BTAD) dataset \cite{mishra2021vt}.} It consists of three distinct categories of industrial components, each characterized by representative body and surface defects. It offers comprehensive annotations at both the image and pixel levels, facilitating a range of anomaly detection tasks. For the three categories, the training sets contain 400, 399, and 1000 defect-free images, respectively. The corresponding test sets comprise 21/49, 30/200, and 410/31 images, denoting the number of normal versus defective samples in each case.

\textbf{Training details.} 
Before being fed into the logical and structural anomaly detection module, all images are resized to $256\times 256$ and $512\times512$ pixels, respectively. Additionally, ResNet101 \cite{he2016deep} and ResNeXt101 \cite{xie2017aggregated} are respectively utilized as the encoder of our logical and structural anomaly detection module. The settings of LeWinBlocks in our Logical Anomaly Detection Module are consistent with \cite{wang2022uformer}, and $\lambda_1$, $\lambda_2$, and $\lambda_3$ are set to 1. Since the global branch is typically more sensitive to logical anomalies and has higher anomaly scores \cite{yao2023learning}, we set the proportion of $\mu$ and $\sigma$ to 1:3 on the MVTec LOCO AD dataset. To train our Logical Anomaly Detection Module, we utilize Adam optimizer \cite{kingma2014adam} with $\beta$ of (0.5, 0.999) and set the learning rate of $0.0001$. Each model is trained for 50 epochs with a batch size of 1 on NVIDIA RTX3090 GPU.

\textbf{Evaluation metrics.} 

To ensure a fair and comprehensive comparison, we adopt widely used evaluation metrics in anomaly detection. Specifically, the area under the receiver operating characteristic curve (AUROC) is used to assess both image-level and pixel-level anomaly localization performance. 
For the MVTec AD dataset, we report the image-level and pixel-level AUROC scores following prior works \cite{cohen2020sub,bergmann2020uninformed,bergmann2022beyond,bergmann2019mvtec}. 
For the MVTec LOCO AD dataset, we report image-level AUROC and the pixel-level saturated per-region overlap (pixel-sPRO), which measures the normalized area under the sPRO curve up to a false positive rate of 0.05, in accordance with \cite{bergmann2022beyond}. 
These metrics quantitatively reflect the model’s ability to distinguish anomalies at both global and local scales.

 \begin{figure}[thbp]
  \centering
  \includegraphics[width=0.5\textwidth]{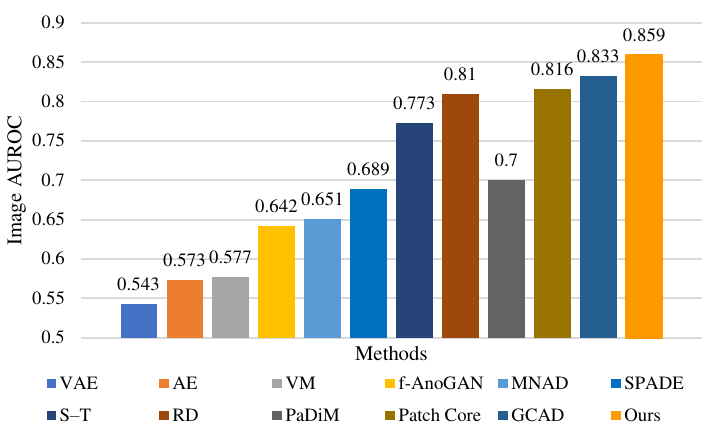}
  \caption{Quantitative results on MVTec LOCO AD dataset for anomaly detection, as measured on the average image-AUROC.} 
   \label{Im}
\vspace{-0.2cm}
\end{figure}

\begin{table*}
 \captionsetup{justification=raggedright,singlelinecheck=false}
 \caption{Quantitative results on the MVTec AD dataset for anomaly detection/localization, as measured on AUROC [\%]. The best results are marked in red, and the second-best results are marked in blue.}
   \resizebox {\textwidth}{!}
   {

       \begin{tabular}{l|lllll|llllllllll|l}
\hline
 \rule{0pt}{11pt} \multirow{2}*{\diagbox{Method}{Category}} & \multicolumn{5}{c|}{Textures}         & \multicolumn{10}{c}{Objects}   \vline & \multirow{2}*{Avg}  \\
\cline{2-16}
  \rule{0pt}{11pt}           & Carpet & Grid  & Leather & Tile  & Wood  & Bottle & Cable & Capsule & Hazelnut & Metal Nut & Pill  & Screw & Toothbrush & Transistor & Zipper &  \\

   \hline

   \rule{0pt}{11pt}   Patch SVDD  \cite{yi2020patch}& 92.9/92.6  & 94.6/96.2  & 90.9/97.4  & 97.8/91.4  & 96.5/90.8  & 98.6/98.1  & 90.3/96.8  & 76.7/95.8  & 92.0/97.5  & 90.9/98.0  & 86.1/95.1  & 81.3/95.7  & 100/98.1  & 91.5/97.0  & 97.9/95.1  & 92.1/95.7  \\
     \rule{0pt}{11pt}   {IGD    \cite{chen2022deep}} & 82.8/94.7 & 97.8/97.7 & 95.8/ {99.5} & 99.1/78.0 & 94.6/89.1 &  {100}/92.2 & 90.6/84.7 & 91.5/97.7 & 99.7/98.0 & 91.3/92.6 & 87.3/97.3 & 82.5/97.0 & 99.7/97.7 & 90.6/84.4 & 97/96.7 & 93.4/93.1 \\

   \rule{0pt}{11pt}   S-T  \cite{bergmann2020uninformed} & 95.3/\ \ -- \ \ \   & 98.7/\ \ -- \ \ \   & 93.4/\ \ -- \ \ \   & 95.8/\ \ -- \ \ \   & 95.5/\ \ -- \ \ \   & 96.7/\ \ -- \ \ \   & 82.3/\ \ -- \ \ \   & 92.8/\ \ -- \ \ \   & 91.4/\ \ -- \ \ \   & 94.0/\ \ -- \ \ \   & 86.7/\ \ -- \ \ \   & 87.4/\ \ -- \ \ \   & 98.6/\ \ -- \ \ \   & 83.6/\ \ -- \ \ \   & 95.8/\ \ -- \ \ \   & 92.5/\ \ -- \ \ \   \\
  \rule{0pt}{11pt}  MKD  \cite{salehi2021multiresolution} &   79.3/95.6    &   78.0/91.8    &   95.0/98.0    &   91.6/82.8    &   94.3/84.8    &   99.4/96.3    &   89.2/82.4    &    80.5/95.9   &  98.4/94.6     &    73.5/86.4   &  82.7/89.6     &   83.3/96.0    &   92.2/96.1    &   85.5/76.5    &   93.2/93.9    & 87.7/90.7  \\

    \rule{0pt}{11pt}  {STFPM} \cite{wang2021student}  & \ -- \ /98.8  & \ -- \ /99.0  & \ -- \ /99.3  & \ -- \ /97.4  & \ -- \ /97.2  & \ -- \ /98.8  & \ -- \ /95.5  & \ -- \ /98.3  & \ -- \ /98.5  & \ -- \ /97.6  & \ -- \ /97.8  & \ -- \ /98.3  & \ -- \ /98.9  & \ -- \ /82.5  & \ -- \ /98.5  & 95.5/97.0  \\
  {  RD     \cite{deng2022anomaly}} & 98.9/ {98.9} &  {100}/99.3 &  {100}/99.4 & 99.3/95.6 & 99.2/95.3 &  {100}/98.7 & 95.0/97.4 & 96.3/98.7 & 99.9/98.9 &  {100}/97.3 & 96.6/ {98.2} & 97.0/99.6 & 99.5/ {99.1} & 96.7/92.5 & 98.5/98.2 & 98.5/97.8 \\

      \rule{0pt}{11pt}       DRAEM  \cite{zavrtanik2021draem} & 97.0/95.5  & 99.9/ {99.7}  &  {100}/98.6  & 99.6/ {99.2}  & 99.1/96.4  & 99.2/ {99.1}  & 91.8/94.7  &  {98.5}/94.3 &  {100}/ {99.7}  & 98.7/ {99.5}  &  {98.9}/97.6  & 93.9/97.6  &  {100}/98.1  & 93.1/90.9  &  {100}/98.8 & 98.0/97.3  \\
  \rule{0pt}{11pt}  CutPaste  \cite{li2021cutpaste} & 93.9/98.3  &  {100}/97.5  &  {100}/ {99.5}  & 94.6/90.5  & 99.1/95.5  & 98.2/97.6  & 81.2/90.0  & 98.2/97.4  & 98.3/97.3  & 99.9/93.1  & 94.9/95.7  & 88.7/96.7  & 99.4/98.1  & 96.1/93.0  & 99.9/ {99.3}  & 96.1/96.0  \\

    \rule{0pt}{11pt} DifferNet  \cite{rudolph2021same} & 92.9/\ \ -- \ \ \   & 84.0/\ \ -- \ \ \   & 97.1/\ \ -- \ \ \   & 99.4/\ \ -- \ \ \   & 99.8/\ \ -- \ \ \   & 99.0/\ \ -- \ \ \   & 95.9/\ \ -- \ \ \   & 86.9/\ \ -- \ \ \   & 99.3/\ \ -- \ \ \   & 96.1/\ \ -- \ \ \   & 88.8/\ \ -- \ \ \   & 96.3/\ \ -- \ \ \   & 98.6/\ \ -- \ \ \   & 91.9/\ \ -- \ \ \   & 95.1/\ \ -- \ \ \   & 94.9/\ \ -- \ \ \   \\

     \rule{0pt}{11pt}  Panda  \cite{reiss2021panda} &    \ -- \ /\ \ -- \ \ \    &    \ -- \ /\ \ -- \ \ \    &    \ -- \ /\ \ -- \ \ \    &   \ -- \ /\ \ -- \ \ \     &     \ -- \ /\ \ -- \ \ \   &    \ -- \ /\ \ -- \ \ \    &   \ -- \ /\ \ -- \ \ \     &    \ -- \ /\ \ -- \ \ \    &    \ -- \ /\ \ -- \ \ \    &     \ -- \ /\ \ -- \ \ \   &    \ -- \ /\ \ -- \ \ \    &   \ -- \ /\ \ -- \ \ \     &    \ -- \ /\ \ -- \ \ \    &    \ -- \ /\ \ -- \ \ \    &    \ -- \ /\ \ -- \ \ \    & 86.5/\ \ -- \ \ \   \\

    \rule{0pt}{11pt}  PaDiM  \cite{defard2021padim} & \ -- \ /99.1  & \ -- \ /97.3  & \ -- \ /99.2  & \ -- \ /94.1  & \ -- \ /94.9  & \ -- \ /98.3  & \ -- \ /96.7  & \ -- \ /98.5  & \ -- \ /98.2  & \ -- \ /97.2  & \ -- \ /95.7  & \ -- \ /98.5  & \ -- \ /98.8  & \ -- \ /97.5  & \ -- \ /98.5  & 97.9/97.5  \\
  \rule{0pt}{11pt} Patch Core  \cite{roth2022towards} & 98.7/99.0  &  {98.2}/98.7  &  {100}/ {99.3}  & 98.7/95.6  & 99.2/95.0  & 100/98.6  & 99.5/98.4  & 98.1/98.8  & 100/98.7  & 100/98.4  & 96.6/97.4  & 98.1/99.4  & 100/98.7  & 100/96.3  & 99.4/ {98.8}  &\textcolor[rgb]{1,0,0}{99.1} /\color{blue}98.1  \\


  \rule{0pt}{11pt} Ours & 99.4/99.2&99.2/98.7&100/99.3&99.8/96.0&98.4/95.6&100/98.8&98.0/98.2&99.5/99.1&100/99.2&100/98.2&94.4/98.1&96.5/99.5&93.9/99.1&99.9/97.0&99.7/99.0& \color{blue}{98.6}/\color{red}{98.3}  \\
   \hline
    \end{tabular}
   }

  \label{tab:01}
\end{table*}

\begin{table}[tbp]
  \centering

  \caption{Quantitative results on MVTec LOCO AD dataset measured by pixel-sPRO at different integration.}
	 \resizebox {\linewidth}{!}
{
    \begin{tabular}{l|lllll}
    \toprule
    Method & L = 0.01  & L = 0.05  & L = 0.1  & L = 0.3  & L = 1.0 \\
    \midrule
    VM \cite{steger2018machine}    & 0.086 & 0.225 & 0.314 & 0.493 & 0.740 \\
    f-AnoGAN \cite{schlegl2019f} & 0.152 & 0.334 & 0.442 & 0.624 & 0.827 \\
    MNAD \cite{park2020learning}  & 0.176 & 0.339 & 0.447 & 0.643 & 0.853 \\
    AE \cite{an2015variational}    & 0.166 & 0.378 & 0.499 & 0.699 & 0.882 \\
    VAE \cite {bergmann2018improving}  & 0.162 & 0.382 & 0.506 & 0.705 & 0.884 \\
    SPADE \cite{cohen2020sub} & 0.225 & 0.451 & 0.587 & 0.790  & 0.927 \\
    S–T \cite{bergmann2020uninformed}   & 0.402 & 0.626 & 0.717 & 0.836 & 0.937 \\
   RD \cite{deng2022anomaly}  & 0.454 & 0.642 & 0.723 & 0.854 & 0.951 \\
 PaDiM \cite{defard2021padim}  & 0.253 & 0.502 & 0.634 &0.821 &0.941 \\
Patch Core  \cite{roth2022towards}  & 0.271 & 0.546 & 0.675 & 0.845 & 0.948 \\
GCAD  \cite{bergmann2022beyond}  & 0.462 & 0.701 & 0.787 & 0.891 & 0.962 \\
    
    Ours & \textbf{0.546} & \textbf{0.758} & \textbf{0.829} & \textbf{0.916} & \textbf{0.972} \\
    \bottomrule
    \end{tabular}%
  \label{tab:l}%
}
\end{table}%


\subsection{Anomaly Detection and Localization}

\textbf{Results on MVTec LOCO AD.}
In the MVTec LOCO AD dataset, all classes contain both logical and structural anomalies with a ratio of approximately 1:1. We compare our proposed method with several state-of-the-art methods, and the quantitative results are summarised in Table \ref{tab:f} and Figure \ref{Im}. The fairness of unsupervised anomaly detection task lies in not introducing prior knowledge of anomalies in the training stage, and the caption annotation (e.g. `breakfast box', `juice bottle', etc.) simply provides names of the image categories, but does not carry any anomaly information. Consequently, our comparative experiment was conducted fairly.

It is apparent that PaDiM and Patch Core rely predominantly on local feature representations to detect anomalies, enabling them to excel at identifying structural defects such as surface cracks, texture irregularities and minor missing components. However, because these approaches lack a comprehensive semantic understanding, they struggle to identify logical anomalies that depend on high-level contextual cues, leaving issues involving mis-assemblies and incorrect object relationships. In contrast, encoder-decoder based methods (e.g., RD and GCAD) are generally more effective in detecting logical anomalies, as they focus on overall reconstruction fidelity.

In our work, we introduce multi-semantics of normality to guide the reconstruction process, integrating both local and global context information derived from normal data distributions. Specifically, we leverage a vector-quantised codebook trained on normal samples to inject fixed semantic priors, and incorporate global semantics from CLIP to introduce high-level conceptual understanding. These components collectively form a semantic bottleneck that biases the model toward reconstructing only normal patterns, thereby suppressing the reproduction of anomalous features.
From a theoretical perspective, this design constrains the decoder’s output space to the learned distribution of normal features. As abnormal patterns are not encoded within the codebook or captured by the global prior, the decoder struggles to reproduce such patterns, leading to higher reconstruction errors. This mechanism functions as a semantic filter, amplifying the discrepancy between normal and abnormal reconstructions, especially for logical anomalies.
Furthermore, we adopt a gradient-preference-based feature selection mechanism and a discriminative feature learning strategy to reinforce anomaly sensitivity during training. This combination ensures that fine-grained structural anomalies are detected while also capturing semantically inconsistent regions effectively.

As a result, our method achieves the best performance on four categories in the pixel-sPRO metric and significantly improves the average pixel-sPRO by 5.7\%, reaching a new SOTA score of 75.8\%. Following previous studies \cite{deng2022anomaly}, Table \ref{tab:l} reports the performance of various comparison methods in pixel-sPRO at different integration intervals (i.e., the maximum false positive rate), clearly demonstrating that our method consistently achieves SOTA performance across these ranges.

Similarly, as shown in Figure \ref{Im}, our method attains a state-of-the-art image-AUROC of 85.9\%, outperforming all competing methods by a substantial margin of 2.6\%. Furthermore, the qualitative results presented in Figure \ref{Vision} illustrate that our approach can accurately localize both logical and structural anomalies.

\begin{table}[htbp]
  \centering
  \caption{Quantitative results on the BTAD dataset for anomaly detection/localization, as measured on AUROC [\%]. The best results are marked in bold.}
 
    \begin{tabular}{l|cccc}
  \hline
     \multirow{2}*{\diagbox{Method}{Dataset} }& \multicolumn{3}{c}{BTAD} & \multirow{2}{*}{AVG}  \\
\cline{2-4}
        & 1     & 2     & 3          \\
   \hline
    Patch-SVDD  \cite{yi2020patch} & 95.7/92.3 & 74.6/93.7 & 82.8/90.8 &  84.4/92.2  \\
    IGD   \cite{chen2022deep}    & 92.3/79.9 & 63.5/83.9 & 91.7/90.3  & 82.5/84.7   \\
    S-T  \cite{bergmann2020uninformed}  & 93.4/83.6 & 88.8/95.4 & 98.7/81.8 &  93.6/81.8 \\
    MKD    \cite{salehi2021multiresolution} & {93.6}/85.4  & {75.6}/86.1  & \textbf{100}/99.1  &  89.8/90.2  \\
    STFPM \cite{wang2021student}   & 90.5/94.7 & 81.2/97.3 & 99.3/99.4 &  90.3/97.1  \\
    RD\cite{deng2022anomaly}& 98.1/96.4 & 82.1/\textbf{96.6} &\textbf{100/99.7} &  93.4/97.5  \\
    DRAEM \cite{zavrtanik2021draem}  & 78.6/68.1 & 80.5/89.1 & 99.5/87.1 & 86.2/81.4   \\
    CutPaste   \cite{li2021cutpaste} &  {95.9}/\ \ -- \ \ &  {85.1}/\ \ -- \ \ & {98.9}/\ \ -- \ \ & {  93.3}/\ \ -- \ \  \ \ \\
    DifferNet \cite{rudolph2021same}  &  {99.1}/\ \ -- \ \ &  {83.9}/\ \ -- \ \ & {97.4}/\ \ -- \ \ & {  93.5}/\ \ -- \ \  \ \ \\
    Panda \cite{reiss2021panda}  & 96.4/96.4 & 82.2/95.9 & 99.9/99.3 & 92.8/97.2  \\
    PaDiM   \cite{defard2021padim}  & 99.4/\textbf{97.2} & 82.5/95.2 & 99.9/99.7 & 93.9/97.4   \\
    Patch Core  \cite{roth2022towards}  & 98.0/96.9 & 81.6/95.8 & 99.8/99.1 & 93.1/\textbf{97.3} \\
    Ours   & \textbf{99.5}/97.0 & \textbf{85.6}/95.1 & \textbf{100}/99.5 & \textbf{95.1}/97.2  \\
   \hline
    \end{tabular}%
  \label{tab:BTAD}%
\end{table}%


\textbf{Results on MVTec AD.}

To demonstrate the superiority of our proposed method, we compare it with state-of-the-art techniques on a classic dataset that predominantly exhibits structural anomalies, with the quantitative results provided in Table \ref{tab:01}. Our method achieves competitive performance in both anomaly detection (with an image-level AUROC of 98.6\%) and localization (with a pixel-AUROC of 98.3\%).

Notably, although our method yields an image-level AUROC that is 0.5\% lower than PatchCore on the MVTec AD dataset, it demonstrates substantial superiority on the MVTec LOCO AD dataset—achieving a 21.2\% improvement in pixel-sPRO and a 2.6\% increase in image-level AUROC. These performance gains are attributed to the key architectural innovations of our method.
Specifically, our gradient-preference-based feature selection mechanism prioritizes high-gradient regions that are more likely to correspond to anomalous patterns, thereby suppressing interference from redundant or irrelevant features. Furthermore, by leveraging multi-semantic features extracted from normal data to guide the reconstruction process, the model effectively captures both fine-grained structural anomalies and high-level logical inconsistencies. This synergy between selective feature emphasis and semantic-guided reconstruction enhances both detection accuracy and anomaly localization performance.

\begin{table}[t]
  \centering
  \caption{Ablation studies on different datasets of our method. }
 \resizebox {\linewidth}{!}{
    \begin{tabular}{ll|ccc}
    \hline
    \multicolumn{2}{l|}{} & \makecell{MVTec LOCO AD\\(pixel-sPRO)}  & \makecell{MVTec AD \\(pixel/image AUROC)}\\
    \hline
   \multicolumn{1}{l|}{\multirow{4}[2]{*}{Logical Branch}} & baseline & 0.596 & 0.961 / 0.916 \\
    \multicolumn{1}{l|}{} & baseline+AGU & 0.605 & 0.969 / 0.930 \\
    \multicolumn{1}{l|}{} & baseline+NMC & 0.609 & 0.968 / 0.926 \\
    \multicolumn{1}{l|}{} & baseline+NMC+AGU & 0.622 & 0.971 / 0.958 \\
    \hline
   \multicolumn{2}{l|}{Structural Branch} & 0.680  & 0.983 / 0.977 \\
    \hline
    \multicolumn{2}{l|}{Structural Branch+Logical Branch} & 0.758 & 0.983 / 0.986  \\
    \hline
    \end{tabular}%
}
  \label{tab:Aa}%

\end{table}%

\subsection{Results on BTAD}

Table \ref{tab:BTAD} reports the image‑level and pixel‑level AUROC scores of various anomaly detection methods evaluated on the BTAD dataset. Our model achieves the highest average image‑level AUROC of 95.1\%, exceeding the previous best performance, while maintaining a competitive average pixel‑level AUROC of 97.2\%. This demonstrates that our approach excels both at identifying defective samples and at accurately localizing defect regions.

For the first object category, our method achieves an image-level AUROC of 99.5\%, demonstrating exceptional detection accuracy. The corresponding pixel-level AUROC of 97.0\% indicates strong capability in localizing small-scale surface irregularities. In the second category, which involves more complex structural anomalies, our model obtains an image-level AUROC of 85.6\%, representing a notable improvement over existing multi-scale and teacher-student frameworks. A pixel-level AUROC of 95.1\% further highlights its robustness in capturing varied anomaly shapes with consistent boundary delineation.

In the third category, near-perfect detection performance is observed across all methods. Our approach reaches an image-level AUROC of 100\%, matching the best reported results. Beyond absolute accuracy, our framework exhibits the smallest gap between detection (image-level) and localization (pixel-level) metrics, indicating a well-balanced optimization of global discrimination and local precision. In contrast, other methods show more pronounced discrepancies between these metrics, reflecting inherent trade-offs between detecting anomalies at the image level and accurately localizing them at the pixel level.

\subsection{Discussion}

\textbf{Ablation analysis.}
We perform an ablation study on the MVTec LOCO AD and MVTec AD datasets to assess the impact of various semantic features and two branches. As reported in Table \ref{tab:Aa}, where baseline, +Global Context Upsampling (AGU), and +Normal Multi-level Codebook (NMC) indicate that utilizing one-class features, +abstract global context features, and +different-level semantic features to reconstruct input images, respectively. The performance has been improved when introducing abstract global context features or different-level semantic features. The inclusion of abstract
global context features allows for a more holistic understanding of the input, while different-level semantic features capture fine-grained details. By leveraging both types of features, our method achieves enhanced performance in anomaly detection tasks. Furthermore, for better evaluation, the MVTec LOCO AD dataset is divided into two subsets, each exclusively containing samples of logical and structural anomalies from all classes, respectively.  Thus, we further conduct experiments of our two branches on the two subsets, and the results are shown in Table \ref{tab:add}. Our logical and structural branches respectively, achieve the best performance on the corresponding subsets. Our logical anomaly detection module achieves 69.0\% pixel-sPRO on the subset of logical anomalies, improving 14.9\% over that of our structural anomaly detection module. On the subset of structural anomalies, the structural anomaly detection module achieves 82.0\% pixel-sPRO, which is 26.7\% higher than the performance of the logical detection module. These results prove that both detection modules achieve good performance in their respective areas of expertise.

\begin{table}[t]
  \centering
  \caption{Ablation experiments of two branches on the MVTec LOCO AD dataset}
 \resizebox {\linewidth}{!}
{
    \begin{tabular}{l|cc}
     \hline
    Methods  & \multicolumn{1}{l}{Logical anomalies } & \multicolumn{1}{l}{Structural anomalies } \\
     \hline
    Logical detection module  & 0.690  & 0.553 \\
    Structural detection module  & 0.541 & 0.820 \\
     \hline
    \end{tabular}%
}
  \label{tab:add}%
\end{table}%

\begin{table}[htbp]
	\centering
	\caption{Computational Efficiency analysis experiments on the MVTec LOCO AD dataset}
	\resizebox {\linewidth}{!}{
		\begin{tabular}{lcccc}
			\hline
			Method & Pixel-sPRO   & {FLOPs[$\times10^9$]}&Latency[ms]& Throughput[img/s]\\
			\hline
			S–T \cite{bergmann2020uninformed} & 0.626   & 4468&88&13 \\
			
			PatchCore \cite{roth2022towards} & 0.546  & 159 + kNN&45&52 \\
			GCAD \cite{bergmann2022beyond} & 0.701    & 416&16&82 \\
		


			Ours  & 0.758   & 125+kNN & 21 & 47 \\
			\hline
		\end{tabular}%
	}
	\label{tab:cf}%
\end{table}%

\subsection{Computational Efficiency Analysis}

To evaluate our method’s suitability for real‐time deployment, we benchmark its computational demands and runtime against three representative baselines—S–T, PatchCore, and GCAD—on the MVTec LOCO AD dataset. The results are shown in Table \ref{tab:cf}. Our approach attains a Pixel‐sPRO of 0.758 while incurring only 21\, ms per image, corresponding to a throughput of approximately 47\,images/s.

Our pipeline consists of two parallel branches: the Structural Anomaly Detection Module and the Logical Anomaly Detection Module. The Structural Anomaly Detection Module computes patch‐level structural anomaly scores via a simple distance‐based metric (sPRO) against a pre‐stored reference set. Importantly, this module requires no online network training: all structural scores can be precomputed offline and cached, thereby eliminating any additional per‐image cost at inference.
Consequently, the end‐to‐end latency is entirely dominated by the Logical Anomaly Detection Module, which executes a single forward pass of our convolutional network. This pass entails approximately 125\, GFLOPs and produces semantically informed anomaly scores in 21\, ms per image.

In real-time application scenarios, the two branches are executed concurrently on separate threads. The runtime complexity of the Structural Anomaly Detection Module depends solely on the size of the reference dataset, mirroring the k‐nearest neighbors search cost employed.

\section{Conclusion}
Our work introduces a novel normality prior-guided multi-semantic fusion network for unsupervised anomaly detection. This framework effectively addresses both structural and logical anomaly detection in an unsupervised setting. By extracting and compressing multi‐level features and using a CLIP‐based semantic codebook to steer reconstruction toward normal patterns, our method suppresses anomalous influences and amplifies the distinction between normal and abnormal instances.  Extensive experiments convincingly demonstrate that our method significantly outperforms previous approaches in both anomaly detection and localization.
Despite these promising results, our study also highlights a critical challenge: the current detection strategies for logical and structural anomalies remain relatively disjoint. While our method integrates mechanisms to handle both types, the underlying modeling and optimization objectives are still partially decoupled. In future work, we aim to explore unified representations that can jointly encode both structural integrity and semantic consistency. Additionally, developing benchmark datasets that systematically include both logical and structural anomalies in a controlled manner could further facilitate the progress toward a holistic and generalized anomaly detection framework.


\bibliographystyle{IEEEtran}
\bibliography{sample}

\end{document}